\documentclass[lettersize,journal]{IEEEtran}
\usepackage{amsmath,amsfonts}
\usepackage{algorithmic}
\usepackage{array}
\usepackage[caption=false,font=normalsize,labelfont=sf,textfont=sf]{subfig}
\usepackage{textcomp}
\usepackage{stfloats}
\usepackage{url}  
\usepackage{verbatim}       
\usepackage{graphicx} 
\def\BibTeX{{\rm B\kern-.05em{\sc i\kern-.025em b}\kern-.08em
    T\kern-.1667em\lower.7ex\hbox{E}\kern-.125emX}}
\usepackage{balance}

\usepackage{epsfig}                                                       

\usepackage{booktabs}

\usepackage{xcolor}

\DeclareMathOperator*{\argmax}{argmax}
\DeclareMathOperator*{\max2}{max}

\begin{document}                   





\title{{\color{black}Improving EO Foundation Models with Confidence Assessment for enhanced Semantic segmentation}}




\author{Nikolaos Dionelis, Nicolas Longépé
\thanks{Manuscript created May, 2024; received June 11, 2024; revised September 5, 2024. Nikolaos Dionelis and Nicolas Longépé are with the European Space Agency (ESA), $\Phi$-lab, ESRIN, Italy. E-mail: Nikolaos.Dionelis@esa.int.}}








\markboth{IEEE GEOSCIENCE AND REMOTE SENSING LETTERS,~VOL.~XX, NO.~X, MONTH~YEAR}%
{How to Use the IEEEtran \LaTeX \ Templates}

\maketitle                                                                                           

\begin{abstract}          
Confidence assessments of semantic segmentation algorithms are important. Ideally, {\color{black} models should have the ability to predict in advance whether their output is likely to be incorrect. Assessing the confidence levels of model predictions in Earth Observation (EO) classification is essential, as it can enhance semantic segmentation performance and help prevent further exploitation of the results in case of erroneous prediction. The model we developed, Confidence Assessment for enhanced Semantic segmentation (CAS), evaluates confidence at both the segment and pixel levels, providing both labels and confidence scores as output. Our model, CAS, identifies segments with incorrect predicted labels using the proposed combined confidence metric, refines the model, and enhances its performance. This work has significant applications, particularly in evaluating EO Foundation Models on semantic segmentation downstream tasks, such as} land cover classification using Sentinel-2 satellite data. {\color{black}The} evaluation results show that this strategy is effective and that the proposed model CAS outperforms other baseline models.
\end{abstract}    


\section{Introduction}                                                                 
\IEEEPARstart{C}{onfidence} assessment of classification algorithms is important as it is a desirable property of models in real-life applications to \textit{a priori} know if they produce incorrect outputs \cite{ESAsummary}.    
Confidence is a metric between $0$ and $1$ that is a proxy for the probability of correct classification.           
Assigning an accurate calibrated confidence metric to every output of a model is crucial for ensuring its reliability and trustworthiness \cite{ESAsummary, VALUES2024}.
Developing methods that flag outputs where the model should not be trusted (i.e., when the model knows it does not know) is crucial for integrating AI models into operational settings, as {\color{black}highlighted in Earth Systems Predictability (ESP) \cite{ESAFDL}.          
ESP  is a strategy to integrate Earth Observation (EO) and a broad range of predictive models into decision-making.   
ESP is enabled by revolutions in EO, rapid AI predictions, robust confidence quantification, and simulations and digital twins, allowing everyone to learn and make decisions.}   
Therefore, AI models should have the ability to \textit{abstain} in certain cases.
{\color{black}The} main application we examine in this paper is EO Foundation Models \cite{PhilEO2023} {\color{black}and} their evaluation {\color{black}on} semantic segmentation land cover classification on satellite EO Sentinel-2 data \cite{PhilEOEGU, nasa2023}.
In this work, we propose a method to detect segments with incorrect predicted labels and then refine the model to improve its performance and generalization by addressing the identified weak points in both the available data and the model.
{\color{black}Using 
the model we developed, Confidence Assessment for enhanced Semantic segmentation (CAS),}   
we can mitigate the negative effects of incorrect classification, improving model decision-making and also preventing high error rates during inference.
{\color{black}Our contributions include: 1) a proposed combined confidence metric, which considers pixel correlations within segments and integrates various statistics; 2) the main application of evaluating EO Foundation Models with confidence assessment; and 3) the evaluation results of 
CAS
at both the \textit{segment} and pixel levels presented in Sec.~\ref{sec:evalsec}.     
Unlike other approaches, we also assess the correlation between the CAS segment confidence and Intersection over Union (IoU) for land cover mapping.}         

\section{Related Work}                                                                                                                                                               
Detecting actual low confidence predictions in EO semantic segmentation is challenging due to frequent distribution changes and the tendency of deep neural network models to be overconfident \cite{VALUES2024, Predictionerrormetaclassification}.   
By using a measure of confidence on the model predictions, we can detect incorrect classifications \cite{ESAsummary, DeVries} and domain shifts.          
{\color{black}A prediction} by the model with high {\color{black}confidence indicates} high reliability and \textit{trust} for this prediction.               
{\color{black}When} a prediction has low confidence, then the model might choose to \textit{abstain} from providing an answer \cite{ESAFDL, newUQmain}. 
{\color{black}Many} EO examples of low confidence \textit{segment} detection exist \cite{DLRGawlikowski}:  
identifying 
big geographical variations, e.g. forests in Europe versus \textit{Africa}, and buildings in Europe versus Asia, for domain adaptation. 
This involves focusing on a specific set of classes where high accuracy is desired, e.g. for urban classes.          
Models should be able to operate in \textit{open}-set settings (rather than closed-set environments) and provide predictions along with a confidence metric \cite{ESAsummary}. 
Confidence assessment also helps detect different biomes, as well as multi-sensor differences.

\begin{figure*}[tb]                                    
\centering \begin{minipage}[b]{.1\linewidth}              
  \centering                                                                  
  \centerline{\epsfig{figure=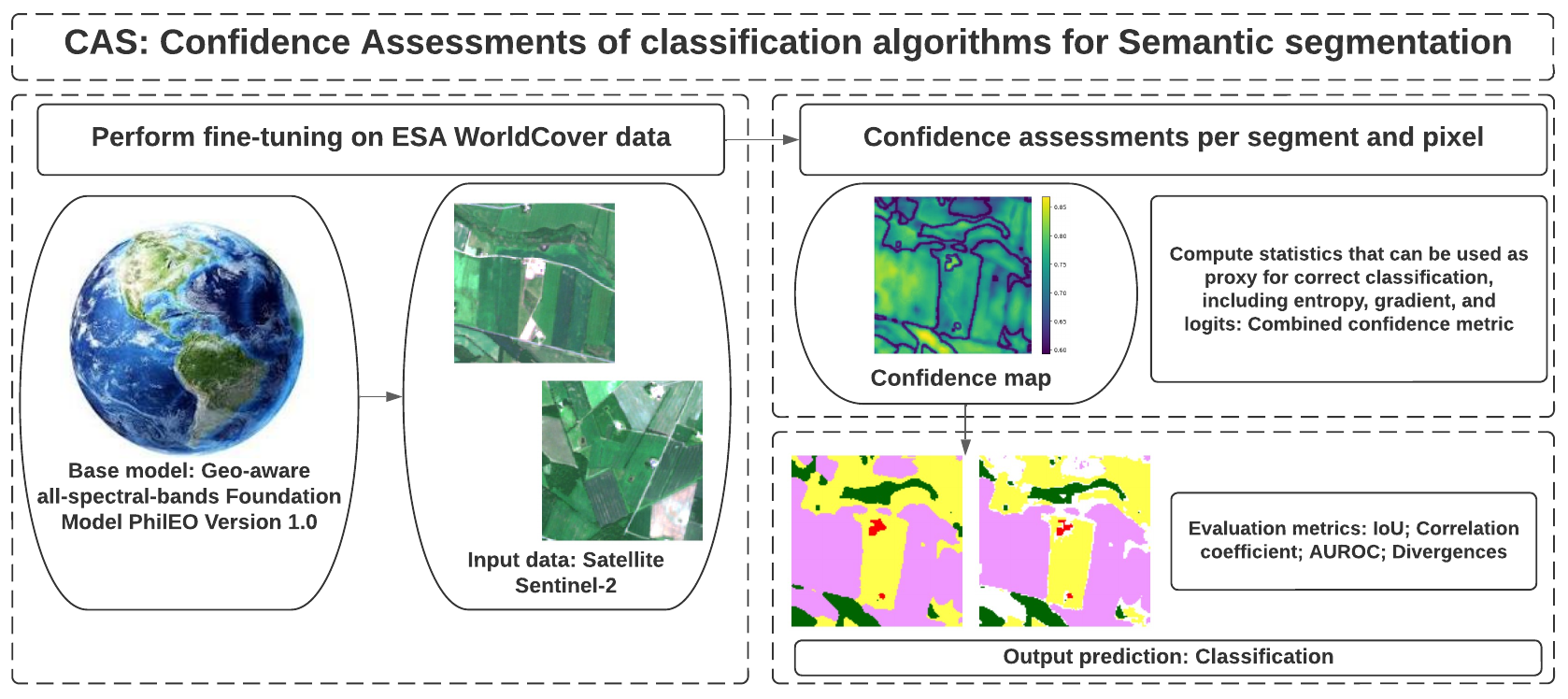,width=12.1cm}}                
  \vspace{0.01cm}  
\end{minipage}     
\vspace{-9pt}        
\caption{\small CAS for \textit{confidence}-aware segmentation: assign confidence metric to predictions, identify wrong predicted labels and refine {\color{black}the model.}}                  
\label{fig:flowchart}                                                     
\end{figure*}

To learn from \textit{fewer} labels, both Foundation and pre-trained models have been trained on unlabeled EO data and evaluated on various applications.         
However, these models do not assign confidence at the segment level for their classification and segmentation predictions \cite{Rottmann2019, Rottmann2021, DetectOoD2017}.  
This lack of confidence assessment and segment-level validation limits their practical adoption, as \textit{end-users} require reliable confidence metrics to trust and effectively integrate them into operational settings.

\section{Proposed Methodology}\label{sec:sectionProposedMethod}                                                                                                          

The proposed model CAS extracts features from {\color{black}EO} data, predicts labels for each pixel and estimates a confidence metric per both segment and pixel.         
We find {\color{black}connected} components, estimate pixel-wise confidence and assign a confidence metric to each segment.           
{\color{black}\text{The flowchart} is shown in Fig.~\ref{fig:flowchart}.}   
CAS computes several \textit{statistics} for the segments and the pixels within {\color{black}them}.      
These features are a \textit{proxy} for correct classification.  

We use the softmax output probability and compute the difference in pixel-wise probability between the \textit{first} and second predicted classes and the negative entropy over the predicted classes, where {\color{black}these} three measures behave in a \textit{similar} manner.        
{\color{black}We} denote the data by $\textbf{x}$, where for each pixel $z$ of each image $x$, the \textit{label} of this pixel in the semantic segmentation task is denoted by $y$.              
The label is from a fixed set of classes, i.e. $y \in \mathcal{C}=\{ y_1, y_2, \dots, y_q \}$, for example $q=11$ for the dataset ESA WorldCover\footnote{\url{http://worldcover2020.esa.int/data/docs/WorldCover_PUM_V1.1.pdf}}.   
We also denote the weights of the classification model $f(\cdot)$, which has a softmax output layer, by $\textbf{w}$.        
The predicted class label, $\hat{y}_z(x, \textbf{w})$, is given by  
\begin{equation}        
\hat{y}_z(x, \textbf{w}) = \argmax_{y \in \mathcal{C}} \ f_z(y \, | \, x, \textbf{w})\text{,} \label{eq:equationnumber1}         
\end{equation}
where $f_z(y \, | \, x, \textbf{w})$ is a probability distribution on the classes, for the pixel $z$.        
Each is non-negative and between $0$ and $1$, and the sum of the probabilities is $1$.     
The difference in pixel-wise probability between the \textit{first} and second predicted classes is
\begin{equation}         
D_z(x, \textbf{w}) = \max2_{y \in \mathcal{C}} \ f_z(y \, | \, x, \textbf{w}) \ - \max2_{y \in \mathcal{C} \, \backslash \, \hat{y}_z(x, \textbf{w})} \ f_z(y \, | \, x, \textbf{w})\text{,}   \label{eq:equationnumber2} 
\end{equation}
and the negative entropy over the predicted classes is      
\begin{equation}      
E_z(x, \textbf{w}) = \sum_{y \in \mathcal{C}} f_z(y \, | \, x, \textbf{w}) \, \log f_z(y \, | \, x, \textbf{w})\text{.} \label{eq:equationnumber3}            
\end{equation}
{\color{black}When the model is confident} in its output, $D_z$ is high, while for near classes, when the model is \textit{not} confident in its prediction, $D_z$ is low.    
We use the difference between the first two classes, and a similar behaviour (with less strength) also happens for the difference between the first and third categories.   
Also, the entropy is \textit{high} for the uninformed uniform distribution, while for a clear/ distinct first class prediction, the entropy is low.

In addition, we compute the gradient statistic, given by        
\begin{equation}                  
G_z(x, \textbf{w}) = \sum_{y \in \mathcal{C}} f_z(y \, | \, x, \textbf{w}) \, (1-\hat{y}') \, \nabla_{\textbf{w}} f_{z0}(y \, | \, x, \textbf{w})\text{,} \label{eq:equationnumb4}                         
\end{equation}
where the first term is the softmax output probability of class $y \in \mathcal{C}$ (e.g., $y_1$ or $y_2$), $\hat{y}'$ is the \textit{one-hot} encoded predicted class label, which means $0$ for $y \in \mathcal{C}$ not equal to the estimated class and $1$ otherwise, and $f_{z0}(y \, | \, x, \textbf{w})$ is the model without the output softmax layer.              
In addition, in \eqref{eq:equationnumb4}, the last term of the equation is the gradient of the model parameters, $\nabla_{\textbf{w}}$.

We also denote the set of connected components for the image $x$ by $\hat{\mathcal{K}}_x$, where these segments depend on the \textit{estimated} semantic segmentation $\hat{S}_x = \{ \hat{y}_z(x, \textbf{w}) \, | \, z \in x\}$.      
In the same way, we denote the ground truth quantities by $\mathcal{K}_x$ and $S_x$, respectively.          
For each segment in $\hat{\mathcal{K}}_x$, i.e. $k \in \hat{\mathcal{K}}_x$, to model cross-pixel correlations within the segment (rather than only pixel-wise), we compute statistics that are a proxy for (and \textit{indicators} of) correct classification, including the coverage of the pixels with confidence \textit{higher} than $\eta = 90\%$ within the segment.            
This spread of high-probability pixels is given by        
\begin{equation}           
C_k(x, \textbf{w}) = \{ \text{Percentage of  } \hat{y}_z(x, \textbf{w})>\eta \, | \, z \in k\}\text{.} \label{eq:equationnumbbeerr}                
\end{equation}

We also use the \textit{inner} part of the segment, which we denote by $k_i$, where $k_i \subset k$ for each pixel $z$ in the segment that has \textit{all} its adjacent pixels within the segment.            
The segment boundary is denoted by $k_b = k \, \backslash \, k_i$.    
We combine the computed statistics into the proposed confidence metric, and to effectively combine the different features which are for {\color{black} cross-pixel} correlations \cite{Predictionerrormetaclassification, SS2023} and for the pixels within the segment (e.g., logits), we transform the calculated features into a \textit{compatible} form to have comparable values.                              
Using the inner segments $k_i$, because the statistics on the segment boundaries $k_b$ are \textit{low} by definition, and using weights with equal contributions, we add the statistics, including $D_z$, $E_z$, $G_z$ and $C_k$, and this is      
\begin{equation}            
T_z(x, \textbf{w}) = \sigma( D_z + E_z + G_z + C_k ), \text{w/ sigmoid function}\text{.} \label{eq:equationnumbbeerreerr}                
\end{equation}
If this confidence metric has a value \textit{smaller} than a threshold $\tau$, i.e. $T_z < \tau$, for example $\tau=0.2$, then this is a predicted incorrect classification.   
Otherwise, it is predicted correct.   




Using the proposed combined confidence metric, we detect segments and pixels with incorrect predicted labels, and we are able to \textit{refine} any model to improve its segmentation performance and generalization.    
Indeed, with the proposed CAS methodology, the model can abstain from attributing any label or output if the confidence is deemed too low, thereby preventing the use of potentially erroneous predictions.

\section{Evaluation: Experiments and Results} \label{sec:evalsec}                                                                                           

\subsection{Experimentation}      

In this work, we start from the PhilEO model  \cite{PhilEO2023, PhilEOEGU}         
which is a \textit{geo-aware} Foundation Model {\color{black}we have recently developed}{\color{black}. It has been trained} on the global \textit{unlabelled} dataset PhilEO Globe \cite{PhilEOEGU}.        
We start from the \textit{all}-spectral-bands U-Net model{\color{black}, and} as a downstream task, we fine-tune and evaluate our model on the labelled {\color{black}dataset WorldCover using Sentinel-2 (10m res)}.

{\color{black}In many} applications, e.g. crop \textit{yield} estimation, confusing \textit{crops} with grass is an important problem, {\color{black}as} crop yield may be overestimated.         
{\color{black}The} features for the two classes Cropland and Grassland are similar in multi-spectral optical EO data, i.e. green colour.   
{\color{black}This} 
is due to epistemic uncertainty, as it is induced by the lack of detail in the measurement. Epistemic {\color{black}uncertainty is} systematic{\color{black}, caused} by \textit{lack} of knowledge, and it   
{\color{black}can} be reduced 
by learning 
the characteristics of {\color{black}the class} 
using additional {\color{black}information} (e.g., \textit{in-situ} {\color{black}measurements).} 
In addition, aleatoric 
{\color{black}uncertainty is} statistical{\color{black},} related to randomness{\color{black}: the} sample not being a \textit{typical} example of the class.       

\begin{figure}[tb]                                                                                      
\hspace{48pt}      
\begin{minipage}[b]{.1\linewidth}         
  \centering               
  \centerline{\epsfig{figure=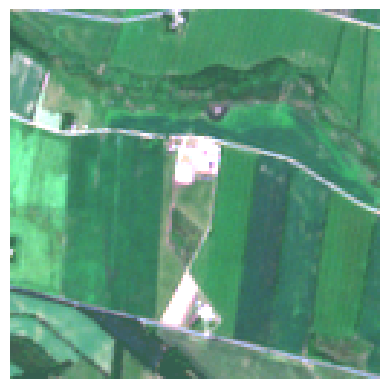,width=3.8cm}}         
  \vspace{-0.01cm}  
  \centerline{\small a) Input image}\medskip 
\end{minipage} 
\hspace{92pt} 
\begin{minipage}[b]{0.1\linewidth} 
  \centering 
  \centerline{\epsfig{figure=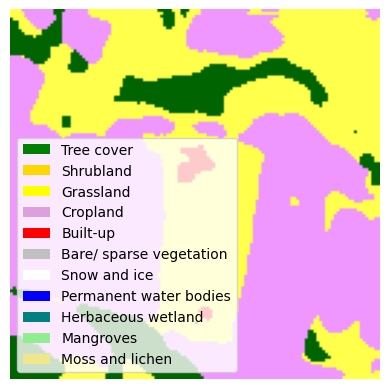,width=3.8cm}}   
  \vspace{-0.01cm}
  \centerline{\small b) Prediction, Base model}\medskip     
\end{minipage}

\hspace{48pt}   
\begin{minipage}[b]{.1\linewidth}           
  \centering      
  \centerline{\epsfig{figure=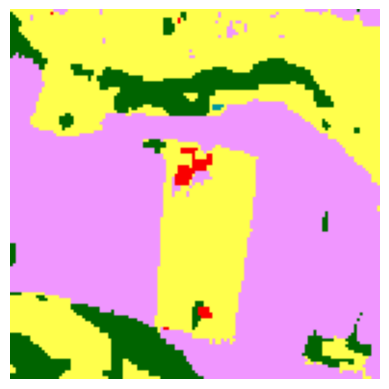,width=3.8cm}}   
  \vspace{-0.01cm}  
  \centerline{\small c) Ground truth}\medskip  
\end{minipage}
\hspace{92pt}
\begin{minipage}[b]{.1\linewidth}           
  \centering      
  \centerline{\epsfig{figure=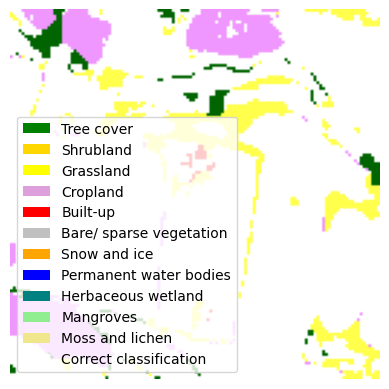,width=3.8cm}} 
  \vspace{-0.01cm}  
  \centerline{\small d) Incorrect classifications}\medskip  
\end{minipage}

\hspace{48pt}   
\begin{minipage}[b]{0.1\linewidth}   
  \centering   
  \centerline{\epsfig{figure=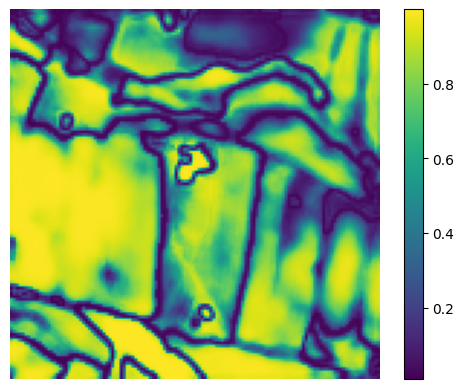,width=4.18cm}}
  \vspace{-0.01cm}
  \centerline{\small e) Confidence map, CAS (Ours)}\medskip
\end{minipage}
\hspace{99pt}       
\begin{minipage}[b]{.1\linewidth}               
  \centering                         
  \centerline{\epsfig{figure=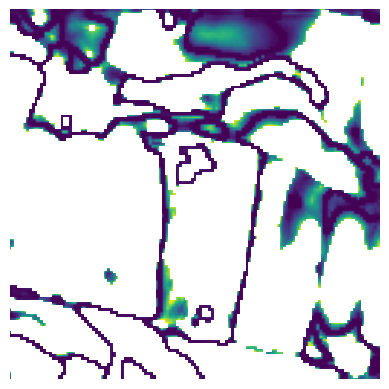,width=3.51cm}}              
  \vspace{-0.01cm}   
  \centerline{\small f) Predicted incorrect, CAS~}\medskip            
\end{minipage}              

\vspace{-9pt}         
\caption{\small Semantic segmentation land cover classification, confidence metric estimation, and confidence assessment by the proposed model CAS on satellite Sentinel-2 data, using the dataset WorldCover. The main classes here are: Tree cover, Grassland, Cropland, and Built-up.}                                                   
\label{fig:figure1CAS}                          
\end{figure}

\begin{figure}[tb]                                                    
\hspace{26pt}    
\begin{minipage}[b]{.1\linewidth}         
  \centering      
  \centerline{\epsfig{figure=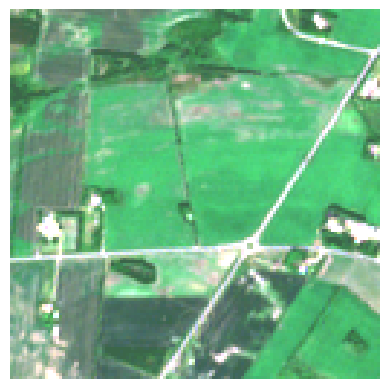,width=2.925cm}}         
  \vspace{0.001cm}   
  \centerline{\small a) Input image~~}\medskip  
\end{minipage}
\hspace{49pt}
\begin{minipage}[b]{0.1\linewidth}
  \centering 
  \centerline{\epsfig{figure=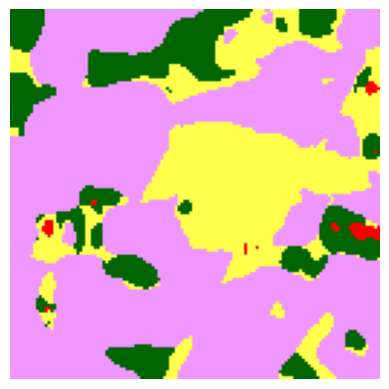,width=2.945cm}}     
  \vspace{0.001cm} 
  \centerline{\small b) Prediction~~}\medskip  
\end{minipage}
\hspace{49pt}    
\begin{minipage}[b]{.1\linewidth}    
  \centering     
  \centerline{\epsfig{figure=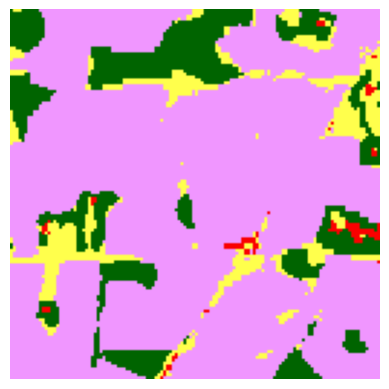,width=2.945cm}}     
  \vspace{0.001cm}   
  \centerline{\small c) Ground truth~~}\medskip 
\end{minipage}              

\hspace{26pt}   
\begin{minipage}[b]{.1\linewidth}          
  \centering         
  \centerline{\epsfig{figure=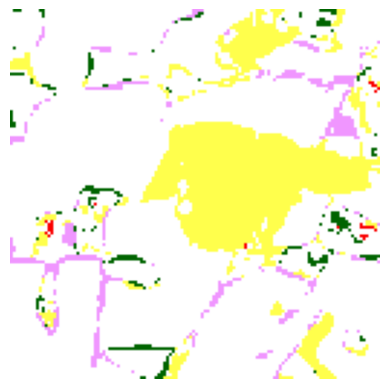,width=2.93cm}}   
  \vspace{0.001cm}  
  \centerline{\small d) Incorrect~~}
  \centerline{\small \, classifications~~}\medskip
\end{minipage}
\hspace{49pt}
\begin{minipage}[b]{0.1\linewidth}   
  \centering         
  \centerline{\epsfig{figure=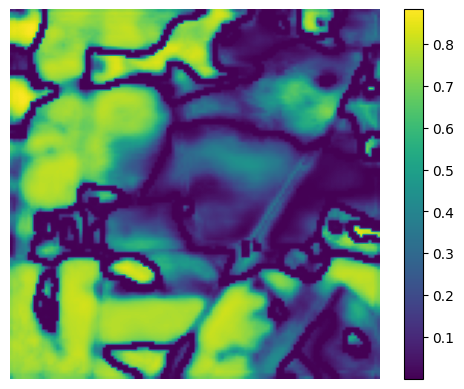,width=2.91cm,height=2.91cm}}       
  \vspace{0.001cm} 
  \centerline{\small e) Confidence~~~}      
  \centerline{\small \, map, CAS (\textit{Ours})~~~}\medskip   
\end{minipage} 
\hspace{49pt}    
\begin{minipage}[b]{.1\linewidth}                
  \centering                       
  \centerline{\epsfig{figure=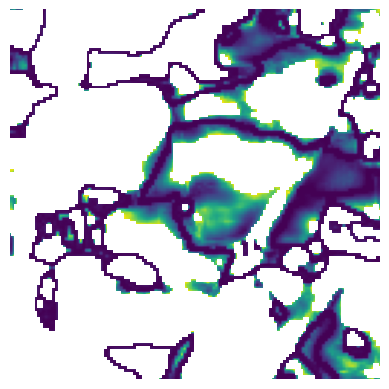,width=2.91cm}}                 
  \vspace{0.001cm}    
  \centerline{\small f) Predicted~}           
  \centerline{\small \, incorrect, CAS~}\medskip        
\end{minipage}    
\vspace{-9pt}        
\caption{\small Semantic segmentation and confidence estimation and assessment by the model CAS described in Sec.~\ref{sec:sectionProposedMethod}, on Sentinel-2 image data using the dataset \textit{ESA WorldCover} with $11$ land cover classes.}                                               
\label{fig:figure4CAS}                             
\label{fig:figure3CAS}   
\end{figure}

\begin{figure}[tb]                                                      
\hspace{26pt}    
\begin{minipage}[b]{.1\linewidth}       
  \centering          
  \centerline{\epsfig{figure=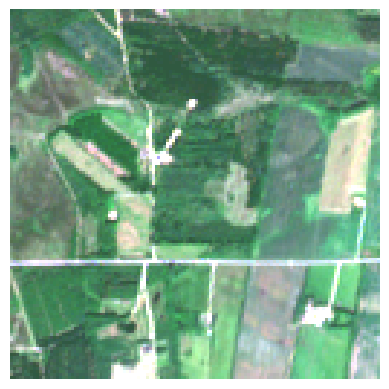,width=2.93cm}}         
  \vspace{0.001cm}  
  \centerline{\small a) Input image~~}\medskip 
\end{minipage}
\hspace{49pt}
\begin{minipage}[b]{0.1\linewidth}  
  \centering   
  \centerline{\epsfig{figure=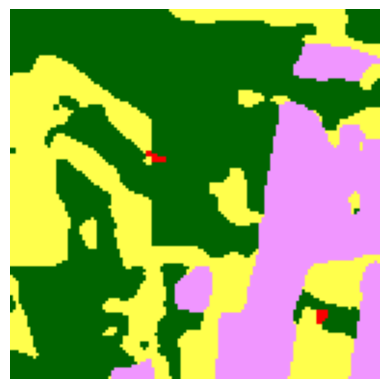,width=2.93cm}}   
  \vspace{0.001cm}
  \centerline{\small b) Prediction~~~}\medskip  
\end{minipage}
\hspace{49pt}    
\begin{minipage}[b]{.1\linewidth}   
  \centering      
  \centerline{\epsfig{figure=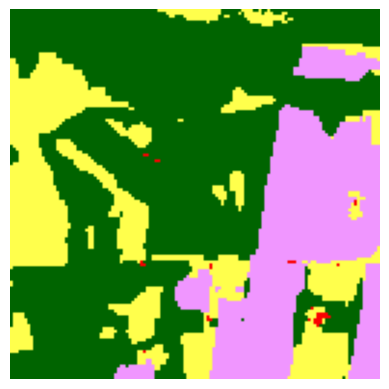,width=2.93cm}}     
  \vspace{0.001cm} 
  \centerline{\small c) Ground truth~~}\medskip 
\end{minipage}              

\hspace{26pt}    
\begin{minipage}[b]{.1\linewidth}              
  \centering              
  \centerline{\epsfig{figure=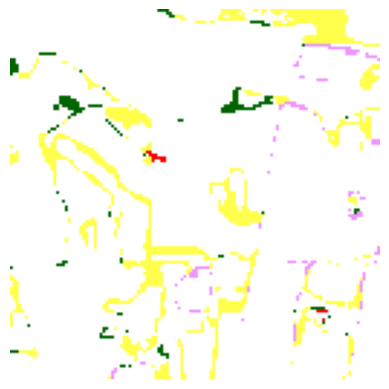,width=2.93cm}}    
  \vspace{0.001cm}  
  \centerline{\small d) Incorrect~~~}
  \centerline{\small \, classifications~~~}\medskip
\end{minipage}
\hspace{49pt}
\begin{minipage}[b]{0.1\linewidth}  
  \centering        
  \centerline{\epsfig{figure=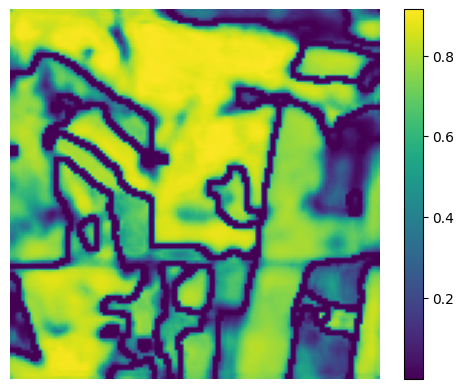,width=2.91cm,height=2.91cm}} 
  \vspace{0.001cm} 
  \centerline{\small e) Confidence~~~~}     
  \centerline{\small \, map, CAS (Ours)~~~~}\medskip   
\end{minipage}
\hspace{49pt}    
\begin{minipage}[b]{.1\linewidth}      
  \centering                   
  \centerline{\epsfig{figure=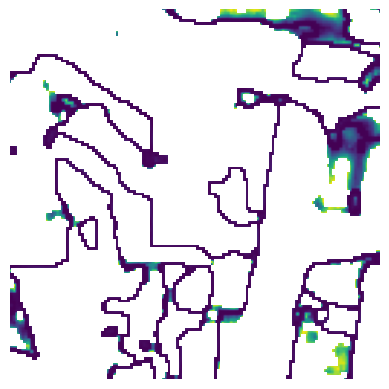,width=2.93cm}}          
  \vspace{0.001cm}  
  \centerline{\small f) Predicted~~}          
  \centerline{\small \, incorrect, CAS~~}\medskip        
\end{minipage}   
\vspace{-9pt}       
\caption{\small Land cover semantic segmentation classification, and both \textit{confidence} assignment and assessment, by the proposed model CAS on Sentinel-2 L2A data trained on the labelled dataset WorldCover.}                                            
\label{fig:figure9CAS}                   
\label{fig:figure5CAS}
\end{figure}

We show the six \textit{images}: {\color{black}a)} Input, b) Prediction, c) Ground truth, d) Incorrect classifications, e) Assigned confidences by CAS, and f) Predicted incorrect classifications, in Figs.~\ref{fig:figure1CAS}-\ref{fig:figure9CAS}.        
We examine the progression of the images chronologically in CAS in (a) to (f) and show the \textit{predicted} incorrect classifications in (f) when the threshold on the confidence metric is $0.2$, i.e. the pixels with a confidence lower than $0.2$ are predicted incorrect. 
The predicted  \textit{a priori} incorrect areas are generally well matching the incorrect classifications based on the ground truths - see (f) and (d). 
Using the assigned confidence by CAS, we find instances where the model simply does not know the correct classification from the available input satellite data.

     

\subsection{Evaluation at segment level}     
We perform \textit{segment-wise} evaluation of the proposed model CAS using both the Intersection over Union (IoU) and the correlation between IoU {\color{black}and} the confidence for the segment.      
The IoU uses the ground truth, as it is based on the \textit{explicit} evaluation of the result.       
On the contrary, the latter, i.e. the confidence for the segment, does not use {\color{black}the} ground truth; it is an \textit{a priori} estimate \cite{VALUES2024, Predictionerrormetaclassification}.       
The \textit{correlation} between the confidence for the segment and the IoU shows the extent {\color{black}to} which the confidence for the segment is a good proxy for the IoU \cite{RottmannM}.                
Also, it is a threshold $\tau$ independent evaluation.  
In the \textit{ideal} case, the correlation is $1$, as high confidence for the predicted segment means correct segmentation: high IoU. 

\textbf{General assessment.}                   
The proposed model CAS achieves an IoU of $74.632\%$ 
in Table~\ref{tab:table1resultsCASmain}, while the base model yields an IoU of $64.282\%$.    
Here, the base model does \textit{not} use confidence assignment, and therefore does not perform segment refinement based on \textit{low confidence} sub-segments (i.e. it does not abstain from providing outputs when \textit{not} confident).     
The percentage improvement of our model CAS compared to the \textit{base} model, i.e. the improvement $((\text{final} - \text{initial}) \, / \, \text{initial})$, is $16.101\%$.     
In addition, the aggregated dispersion measures model from \cite{Rottmann2021} yields an IoU of $69.565\%$.       
Here, the percentage \textit{improvement} of CAS compared to this model is $7.284\%$
in Table~\ref{tab:table1resultsCASmain}.

As further \textit{segment}-wise evaluation of the proposed model CAS, we compute the correlation between the confidence for the segment and  the IoU. CAS achieves the correlation coefficient of $60.529\%$ in Table~\ref{tab:table1resultsCASmain}.  
The correlation when the softmax probability is used on its own, averaged over the segment, i.e. the mean over the \textit{interior} of the segment without its boundary, is $35.735\%$. CAS \textit{improves} the correlation between the confidence for the segment and the IoU. The percentage improvement is $69.383\%$.  
These results demonstrate the superiority of our approach in providing a reliable confidence metric leading to \textit{trustworthy} and more accurate predictions.

\textbf{Sensitivity analysis for CAS.}                   
The above analysis is based on a set of internal parametrizations that can be modified.      
In Table~\ref{tab:table2resultsCASmain2}, different parameters are set, and the corresponding model performances are 
examined.     
When the coverage of the pixels with softmax probability higher than $\eta = 80\%$ (resp. $70\%$) 
in the segment is used (instead of 
$90\%$ probability)   
in Eq. \eqref{eq:equationnumbbeerr}, then the IoU is $73.316\%$ (resp. $73.039\%$).   
We observe that this internal parameter has limited impact on the model performance. However, the analysis for the correlation sensitivity shows an improvement from $60.529\%$ to $63.124\%$ (or resp. $62.645\%$), thus exhibiting a better correlation.

In addition, when the median is used over the segment instead of the average, the performance and reliability of the model to provide an accurate confidence metric, i.e. a proxy for the probability of correct classification, degrades.        
In particular, the IoU \textit{decreases} to $68.325\%$, while the correlation decreases to $57.963\%$.           
Hence, we observe that computing the \textit{mean} over the segment (rather than the median) leads to better results.

Furthermore, when the final refinement to improve the segmentation performance of the model is \textit{not} performed, then the correlation is $46.778\%$ in Table~\ref{tab:table2resultsCASmain2}.      
Therefore, according to the above results, the internal parametrization of $\eta$ is important in the proposed model CAS (i.e. $\eta = 80\%$), while calculating the average over the segment yields the best performance.

\begin{table}[!tb]                                                                    
    \caption{Evaluation of the model CAS in the IoU and correlation metrics on the dataset ESA WorldCover (multi-spectral, $10$ spectral bands). Comparison to the base model, which uses the softmax output probability in \eqref{eq:equationnumber1}, and also to the baseline model \cite{Predictionerrormetaclassification}. In addition, percentage Improvement (I) over the base model we used \cite{PhilEO2023, PhilEOEGU}, i.e. $((\text{final} - \text{initial}) \, / \, \text{initial})$.}\label{tab:table1resultsCASmain}   
      \centering                            
        \begin{tabular} 
{p{1.96cm} p{1.1cm} p{1.1cm} p{1.3cm} p{0.98cm}}   
  
\toprule            
    \normalsize {\small \textbf{Model}} &   
    \normalsize {\small \textbf{IoU ($\%$)}} &    
    \normalsize {\small \textbf{I IoU}} & \normalsize {\small \textbf{Corr. ($\%$)}} & \normalsize {\small \textbf{I Corr.}}       
\\
\midrule               
\midrule  
\normalsize {\small CAS (\textit{Ours}) with Eq. \eqref{eq:equationnumbbeerreerr}}   
& \normalsize {\normalsize  \small $\textbf{74.632}$} &     
    \normalsize {\normalsize  \small $\textbf{16.101}$} & \normalsize {\normalsize \small $\textbf{60.529}$} & \normalsize {\normalsize \small $\textbf{69.383}$}  
\\   
\midrule          
\midrule 
\normalsize {\small Base model \cite{PhilEO2023} with Eq. \eqref{eq:equationnumber1}}   &    
\normalsize {\normalsize  \small $64.282$} & 
\normalsize {\normalsize  \small $N/A$} & \normalsize {\normalsize  \small $\text{35.735}$} & \normalsize {\normalsize  \small $N/A$} 
\\ 
\midrule 
\normalsize {\small Baseline model \cite{Predictionerrormetaclassification,Rottmann2021}} &  
\normalsize {\normalsize  \small $69.565$} &
\normalsize {\normalsize  \small $7.594$} & \normalsize {\normalsize  \small $43.155$} & \normalsize {\normalsize  \small $20.764$}  
\\ 
\midrule                            
\midrule     
\end{tabular}          
\end{table}

\begin{table}[!tb]                                
    \caption{Ablation study of CAS on the dataset ESA WorldCover, and improvement (I) over the model without (w/o) refinement.}\label{tab:table2resultsCASmain2}   
      \centering              
        \begin{tabular}
{p{2.34cm} p{0.85cm} p{0.95cm} p{0.95cm} p{1.04cm}} 
  
\toprule         
    \normalsize {\small \textbf{Model}} & 
    \normalsize {\small \textbf{IoU}} &  
    \normalsize {\small \textbf{I IoU}} & \normalsize {\small \textbf{Corr.}} & \normalsize {\small \textbf{I Corr.}}    
\\
\midrule                  
\midrule 
\normalsize {\small CAS with Eq. \eqref{eq:equationnumbbeerreerr}}       
& \normalsize {\normalsize  \small $\textbf{74.632}$} &   
    \normalsize {\normalsize  \small $\textbf{16.101}$} & \normalsize {\normalsize \small $60.529$} & \normalsize {\normalsize \small $29.396$}   
\\   
\midrule         
\midrule 
\normalsize {\small CAS w/ $\, \eta=80$\%} & 
\normalsize {\normalsize  \small $73.316$} &
\normalsize {\normalsize  \small $14.054$} & \normalsize {\normalsize  \small $\textbf{63.124}$} & \normalsize {\normalsize  \small $\textbf{34.944}$}  
\\ 
\midrule      
\normalsize {\small CAS w/ $\, \eta=70$\%} &  
\normalsize {\normalsize  \small $73.039$} &
\normalsize {\normalsize  \small $13.623$} & \normalsize {\normalsize  \small $62.645$} & \normalsize {\normalsize  \small $33.920$}   
\\ 
\midrule     
\normalsize {\small CAS w/ median} & 
\normalsize {\normalsize  \small $68.325$} &
\normalsize {\normalsize  \small $6.289$} & \normalsize {\normalsize  \small $\text{57.963}$} & \normalsize {\normalsize  \small $\text{23.911}$}  
\\ 
\midrule      
\normalsize {\small CAS w/o refinem.} &     
\normalsize {\normalsize  \small $64.282$} &
\normalsize {\normalsize  \small $N/A$} & \normalsize {\normalsize  \small $46.778$} & \normalsize {\normalsize  \small $N/A$} 
\\ 
\midrule                               
\midrule
\end{tabular}            
\end{table}

\subsection{Evaluation at pixel level}  

We now evaluate the proposed model CAS at the \textit{pixel} level and assess the assigned confidence for all the examined images.     
We compute histograms of the confidence scores for 
the correct classifications and 
misclassifications.      
We calculate distribution distances to 
assess the \textit{separability} of the incorrect and 
correct classifications with the assigned confidence metric.

\begin{figure}[tb]                    
  \centering                     
  \centerline{\epsfig{figure=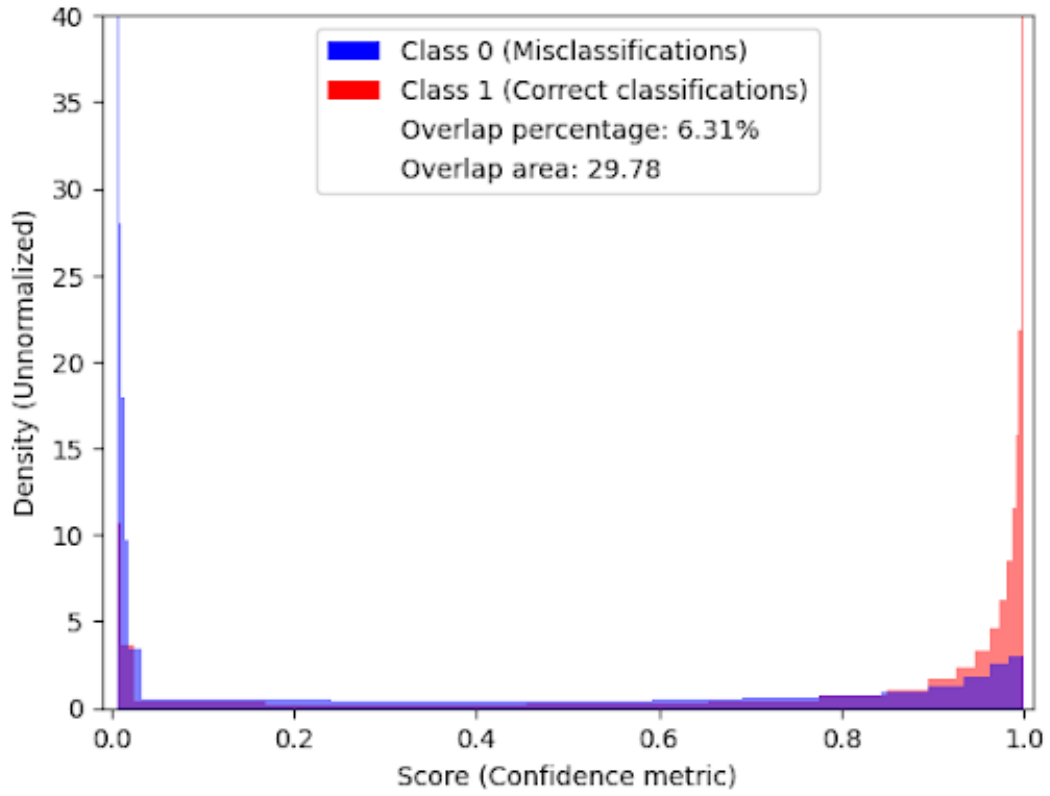,width=8.24cm}} 
\caption{\small Histogram plot for CAS with Eq. \eqref{eq:equationnumbbeerreerr} for \textit{semantic} segmentation land cover classification on Sentinel-2 L2A multi-spectral data using WorldCover. The aim is to effectively \textit{separate} misclassifications and correct classifications. The horizontal axis is the confidence metric.}  
\label{fig:figure6CASnew1} \label{fig:figure6CAS}                                          
\end{figure}

We examine the histograms and the distribution of the scores in Fig.~\ref{fig:figure6CAS}.        
Also, in Table~\ref{tab:table4results}, for the separability of the correct and \textit{incorrect} classifications, we calculate the Kullback-Leibler (KL) and Jensen-Shannon (JS) $f$-divergences, the Wasserstein distance distribution metric, and the threshold $\tau$ independent evaluation metric Area Under the Receiver Operating Characteristics Curve (AUROC).         
In Fig.~\ref{fig:figure6CAS}, the histogram of the proposed model CAS has \textit{two} distinct peaks at $0$ and $1$ for the misclassifications and the correct classifications, respectively. This is desirable.    
The Wasserstein distance is $13.524$, while the JS divergence is $2.805$. 
In addition, the KL divergence is $2.580$ (also $3.029$ as the KL $f$-divergence is non-symmetric) and the AUROC is $0.901$.  
The overlap area percentage is $27.440\%$ for CAS in Table~\ref{tab:table4results}, while the Euclidean distance is $11.987$.

\begin{figure}[tb]                                                                             
\hspace{26pt}    
\begin{minipage}[b]{0.2\linewidth}
  \centering      
  \centerline{\epsfig{figure=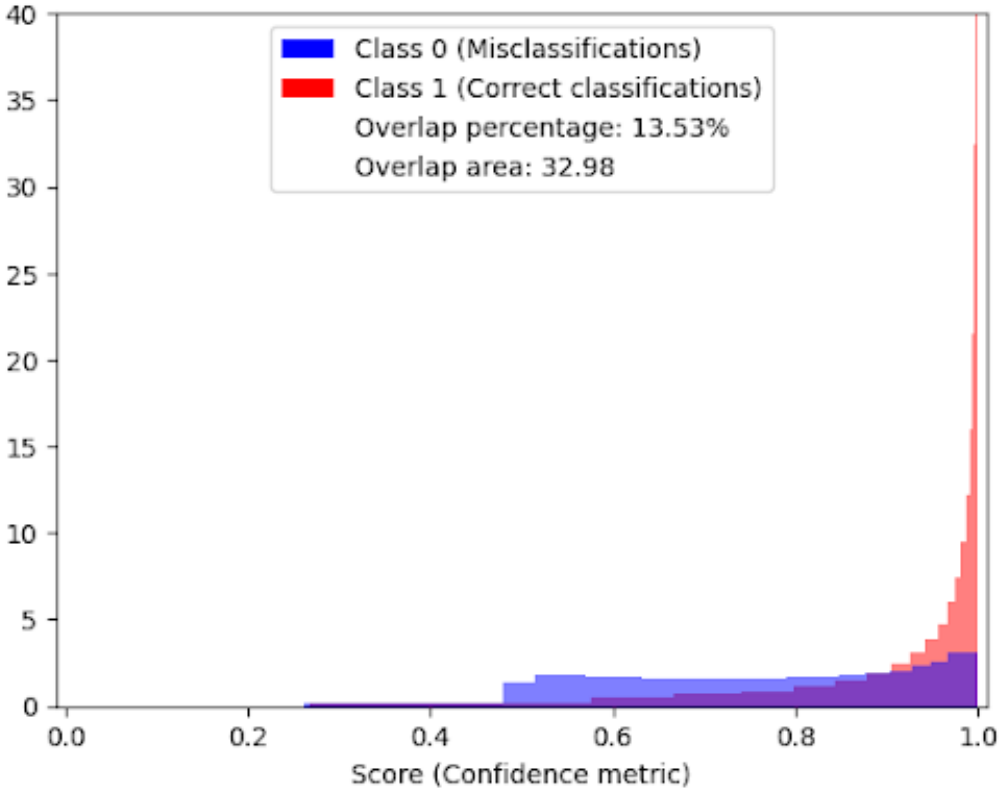,width=4.6cm}}      
  \vspace{0.001cm}
  \centerline{\small \, ~a) Base model \cite{PhilEO2023} with Eq. \eqref{eq:equationnumber1}}\medskip     
\end{minipage}
\hspace{73pt}       
\begin{minipage}[b]{.2\linewidth}            
  \centering                           
  \centerline{\epsfig{figure=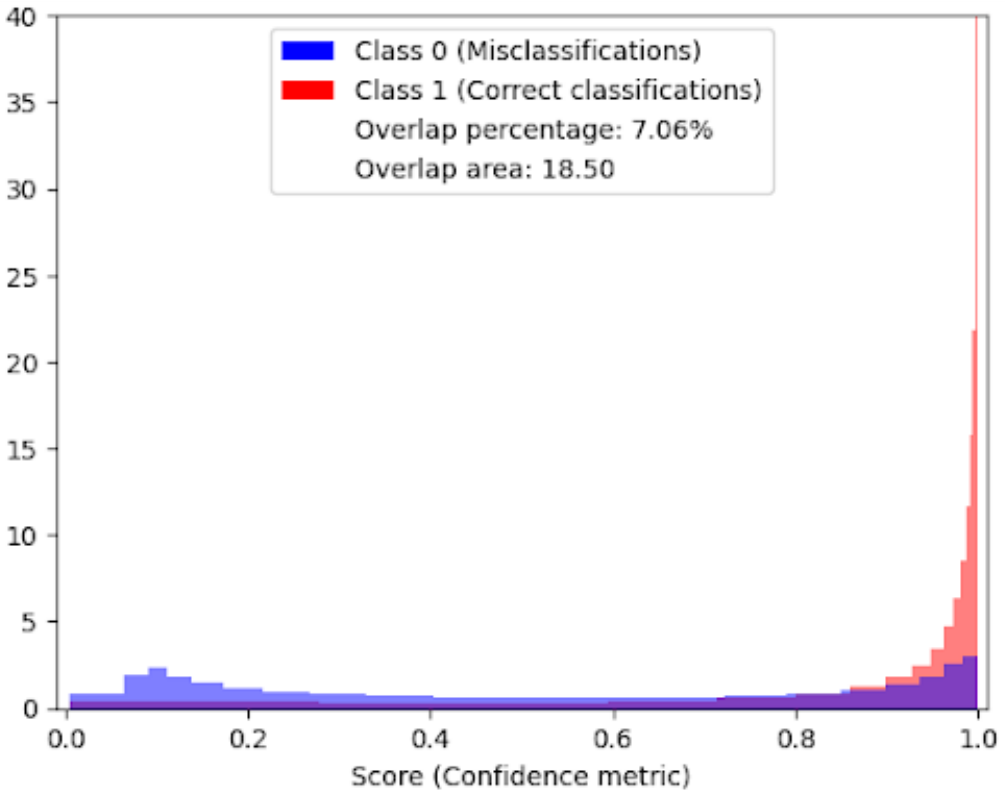,width=4.6cm}}                 
  \vspace{0.001cm}    
  \centerline{\small \, b) Ablation study: w/o Eq. \eqref{eq:equationnumbbeerreerr}}\medskip  
\end{minipage}   
\vspace{-9pt}       
\caption{\small Plots of the histograms for the task of semantic segmentation land cover mapping on Sentinel-2 data using the dataset WorldCover, where the target is to separate incorrect from correct classifications.} 
\label{fig:figure6CASnew2}                                        
\end{figure}

The proposed model CAS outperforms other methods in Fig.~\ref{fig:figure6CASnew2}(a) where the softmax output probability on its own is used.    
For the latter, the Wasserstein distance is $5.071$ in Table~\ref{tab:table4results}.  
The JS divergence is $0.566$, the KL divergence $0.475$ (also $0.658$ since KL is not symmetric), and the AUROC $0.777$.  
Also, here, the overlap area percentage is $33.584\%$, while the Euclidean distance is $9.425$.      
The percentage \textit{improvement} of CAS, compared to when only the softmax probability is used, is $15.959\%$ for the AUROC in Table~\ref{tab:table4results}, and $27.183\%$ for the Euclidean distance.       
We observe that all the evaluation metrics for CAS show \textit{improved} performance compared to the other models.      
Furthermore, as an ablation study, in Table~\ref{tab:table4results}, the model CAS outperforms the model in Fig.~\ref{fig:figure6CASnew2}(b) which does \textit{not} use segment boundaries (comparing Fig.~\ref{fig:figure6CAS} to Fig.~\ref{fig:figure6CASnew2}(b)).

\section{Conclusion}                                                                 
We have proposed the model CAS for confidence assignment and assessment {\color{black}for} semantic segmentation classification tasks.              
CAS takes as input EO satellite Sentinel-2 multi-spectral data, computes confidence and improves the segmentation performance of the model.  
For instances where the model simply does not know the correct classification from the available input data, e.g. due to lack of resolution or spectral information, models should be able to output ``\textit{None} of the above'' for the {\color{black}semantic segmentation output class}.    
The evaluation on the task of land cover classification on the dataset ESA WorldCover shows that CAS outperforms other models in IoU, correlation, JS divergence, AUROC and Wasserstein distance metrics in Tables~\ref{tab:table1resultsCASmain} and \ref{tab:table4results}.   
%
We also release our code for reproducibility\footnote{\url{http://github.com/ESA-PhiLab/CAS_Confidence_Assessment}}.     
As future work, we aim to perform segment-based spectral signature assessment, and we also plan to use the results of CAS for further fine-tuning EO Foundation Models, as well as for noisy labels mitigation to detect incorrect class labels in EO datasets.
In addition, we also plan to use the results of CAS for further improving EO Foundation Models where low confidence areas indicate regions in the data space for: i) collecting new data, or ii) performing specific data augmentation strategies to these samples.  
Assigning a correct calibrated confidence metric to each inference of the model is crucial for models to be robust and operational in real-world scenarios.
Also, accuracy on its own is not a sufficient performance measure for the models and problems we consider. 
We believe that CAS will open the road to confidence assessment of EO Foundation Models and inspire other researchers to adopt this real-world setting.

\begin{table}[!tb]                                                       
    \caption{Evaluation of CAS at the pixel level using distribution metrics, and comparison to base model \cite{PhilEO2023} and ablation study.}\label{tab:table4results}      
      \centering                       
        \begin{tabular}
{p{3.74cm} p{0.8cm} p{1.534cm} p{1.05cm}} 
  
\toprule          
    \normalsize {\small \textbf{Evaluation metrics}} &  
    \normalsize {\small \textbf{CAS}} &  
    \normalsize {\small \textbf{Base model}} & 
    \normalsize {\small \textbf{Ablation}}
\\
\midrule         
\midrule
\normalsize {\small Wasserstein distance $\uparrow$} 
& \normalsize {\normalsize  \small $\textbf{13.524}$} &  
    \normalsize {\normalsize \small $5.071$} &
    \normalsize {\normalsize \small $6.721$}
\\   
\midrule 
\normalsize {\small Jensen-Shannon divergence $\uparrow$} &   
\normalsize {\normalsize  \small $\textbf{2.805}$} & 
\normalsize {\normalsize  \small $0.566$} &
\normalsize {\normalsize \small $0.980$}
\\ 
\midrule   
\normalsize {\small KL divergence $\uparrow$} &  
\normalsize {\normalsize  \small $\textbf{2.580}$} & 
\normalsize {\normalsize  \small $0.475$} &
\normalsize {\normalsize \small $0.630$}
\\ 
\midrule     
\normalsize {\small KL diver. (other way round)} &  
\normalsize {\normalsize  \small $\textbf{3.029}$} &
\normalsize {\normalsize  \small $0.658$} &
\normalsize {\normalsize \small $1.330$}
\\ 
\midrule 
\normalsize {\small AUROC $\uparrow$} &  
\normalsize {\normalsize  \small $\textbf{0.901}$} & 
\normalsize {\normalsize  \small $0.777$} &
\normalsize {\normalsize \small $0.884$}
\\ 
\midrule  
\normalsize {\small Overlap area $\downarrow$} &    
\normalsize {\normalsize  \small $\textbf{27.440}$} &
\normalsize {\normalsize  \small $33.584$} &
\normalsize {\normalsize \small $29.537$}
\\  
\midrule   
\normalsize {\small Euclidean distance $\uparrow$} &  
\normalsize {\normalsize  \small $\textbf{11.987}$} &
\normalsize {\normalsize  \small $9.425$} &
\normalsize {\normalsize \small $10.492$}
\\ 
\midrule                           
\midrule 
\end{tabular}          
\end{table}


\begin{thebibliography}{1}                           


\bibitem{ESAsummary}               
{{European Space Agency (ESA) EO $\Phi$-Week}}, {\it{AI4EO: Recommendations}}, 
p. 19, 
2021. 
Online: \url{https://az659834.vo.msecnd.net/eventsairwesteuprod/production-nikal-public/bbb84824e3564ca2adfde58c4893aa91}    





\bibitem{VALUES2024}      
K. Kahl, C. Luth, et al., 
{\it{VALUES: A Framework for Systematic Validation of Uncertainty Estimation in Semantic Segmentation}}, 
In ICLR, 2024.





\bibitem{ESAFDL}            
{{ESA $\Phi$-lab, University of Oxford, FDL, Trillium}}, {\it{Earth Systems Predictability: How can AI advance planetary stewardship?}}, 
2023. 
Online: \url{http://www.calameo.com/read/005503280e1c3da2978db}







\bibitem{PhilEO2023}          
C. Fibaek, L. Camilleri, A. Luyts, N. Dionelis, and B. Le Saux, 
{\it{PhilEO Bench: Evaluating Geo-Spatial Foundation Models}}, 
IGARSS, 2024.



\bibitem{PhilEOEGU}    
B. Le Saux, et al., 
{\it{The PhilEO Geospatial Foundation Model Suite}}, 
EGU, 2024. http://meetingorganizer.copernicus.org/EGU24/EGU24-17934.html

\bibitem{nasa2023}  
J. Jakubik, et al., 
{\it{FMs for Generalist Geospatial
AI}}, 
arxiv:2310.18660.









\bibitem{SS2023}        
P. de Jorge, et al., 
{\it{Reliability in Semantic Segmentation}}, 
In CVPR, 2023.


\bibitem{Predictionerrormetaclassification}      
M. Rottmann, et al., 
{\it{Prediction error meta classification}}, In IJCNN, 2020.







\bibitem{DLRGawlikowski}     
J. Gawlikowski, et al., 
{\it{An advanced Dirichlet prior for Out-of-Distribution detection in remote sensing}}, 
IEEE TGRS, 2022.








\bibitem{DeVries}   
T. DeVries, 
{\it{Learning Confidence for OoD Detection}}, 
arXiv:1802.04865.





\bibitem{UncertaintySS2024}  
J. Küchler, 
et al., 
{\it{Uncertainty estimates for semantic segmentation: Providing enhanced reliability for automation}}, 
arXiv:2401.09245, 2024.

\bibitem{PixelwiseAD2021} 
G. Di Biase, H. Blum, et al.,
{\it{Pixel-wise Anomaly Detection in Complex Driving Scenes}}, 
In Proc. CVPR, 2021.



\bibitem{Rottmann2019}      
M. Rottmann and M. Schubert, 
{\it{Uncertainty Measures and Prediction Quality Rating for the Semantic Segmentation}}, CVPR Workshop, 2019.

\bibitem{Rottmann2021} 
R. Chan, et al.,
{\it{Entropy Maximization and Meta Classification for Out-of-Distribution Detection in Semantic Segmentation}}, In Proc. ICCV, 2021.

\bibitem{UnmaskingAnomalies2023}     
S. Rai, et al., 
{\it{Unmasking Anomalies in Segmentation}}, In ICCV, 2023.

\bibitem{RottmannM}    
M. Rottmann, 
{\it{Automated detection of label errors in semantic segmentation datasets via deep learning and UQ}}, In Proc. WACV, 2023.




\bibitem{DetectOoD2017}      
D. Hendrycks, 
{\it{Baseline for Detecting Misclassified \& OoD}}, ICLR, 2017.






\bibitem{GeoBench2023}      
A. Lacoste, et al.,
{\it{GEO-Bench: Toward FMs for Earth}}, NeurIPS, 2023.



\bibitem{newUQmain}       
{\color{black}J. Gawlikowski}, et al., {\it{A Survey of Uncertainty in Deep Neural Networks}}, Artif. Intell. Rev. (56), p. 1513-1589, 2023.





\bibitem{newUQmain2}                      
{\color{black}C. Corbière}, et al., {\it{Addressing failure prediction by learning model confidence}}, in Proc. NeurIPS, p. 2898-2909, 2019.





































\end{thebibliography}

\begin{thebibliography}{1}   

\bibitem{ams}
{\it{Mathematics into Type}}, American Mathematical Society. Online available: 

\bibitem{oxford}
T.W. Chaundy, P.R. Barrett and C. Batey, {\it{The Printing of Mathematics}}, Oxford University Press. London, 1954.

\bibitem{lacomp}{\it{The \LaTeX Companion}}, by F. Mittelbach and M. Goossens

\bibitem{mmt}{\it{More Math into LaTeX}}, by G. Gr\"atzer

\bibitem{amstyle}{\it{AMS-StyleGuide-online.pdf,}} published by the American Mathematical Society

\bibitem{Sira3}
H. Sira-Ramirez. ``On the sliding mode control of nonlinear systems,'' \textit{Systems \& Control Letters}, vol. 19, pp. 303--312, 1992.

\bibitem{Levant}
A. Levant. ``Exact differentiation of signals with unbounded higher derivatives,''  in \textit{Proceedings of the 45th IEEE Conference on Decision and Control}, San Diego, California, USA, pp. 5585--5590, 2006.

\bibitem{Cedric}
M. Fliess, C. Join, and H. Sira-Ramirez. ``Non-linear estimation is easy,'' \textit{International Journal of Modelling, Identification and Control}, vol. 4, no. 1, pp. 12--27, 2008.

\bibitem{Ortega}
R. Ortega, A. Astolfi, G. Bastin, and H. Rodriguez. ``Stabilization of food-chain systems using a port-controlled Hamiltonian description,'' in \textit{Proceedings of the American Control Conference}, Chicago, Illinois, USA, pp. 2245--2249, 2000.

\end{thebibliography}
\end{document}